\pdfoutput=1
 
\documentclass[11pt]{article}
 
\usepackage{acl}
 
\usepackage{times}
\usepackage{latexsym}
\usepackage{multirow}

\usepackage[T1]{fontenc}
 
\usepackage[utf8]{inputenc}
 
\usepackage{microtype}
 
\usepackage{inconsolata}
 
\usepackage{graphicx}

\usepackage{float}
 
\title{BLT: Can Large Language Models Handle Basic Legal Text?}
 
\author{Andrew Blair-Stanek$^{1,2}$ \ \ \ \  Nils Holzenberger$^{3}$\ \ \ \ \ Benjamin Van Durme$^{1}$ \\ 
$^{1}$ Johns Hopkins University, \hspace{1ex} $^{2}$ University of Maryland School of Law, \\ 
$^{3}$ T\'el\'ecom Paris - Institut Polytechnique de Paris}
 
\begin{document}
\maketitle
\begin{abstract}
We find that the best publicly available LLMs like GPT-4 and Claude currently perform poorly on basic legal text handling. This motivates the creation of a benchmark consisting of examples that lawyers and paralegals would expect LLMs to handle zero-shot, such as looking up the text at a line of a witness deposition or at a subsection of a contract. LLMs' poor performance on this benchmark casts into doubt their reliability as-is for legal practice. However, fine-tuning on our training set brings even a small model to near-perfect performance.  This benchmark will be useful for fine-tuning LLMs for downstream legal tasks, as well as for tracking LLMs' reliability as-is for basic legal tasks. 
\end{abstract}
 
\section{Introduction}
 
There has been much excitement about the potential of large language models (LLMs) to handle complex reasoning tasks.
In the legal domain, recent work has focused on LLMs' ability to perform high-level tasks such as passing the bar exam,  identifying hearsay evidence, or navigating the tax code.  These 
results have led to predictions that LLMs will imminently revolutionize law practice.

\begin{figure}[!ht]
 
1: MR. MACEJKA: Object to form.
 
2: A. The medical invoice from the MAO would be
 
3: the medical claims data. So we're already in
 
4: receipt of that.
 
5: BY MR. PENNEKAMP:
 
6: Q. That's what you referred to a medical
 
7: invoice, is the claims data?
 
8: MR. MACEJKA: Object to form.
 
9: A. That essentially is what the medical
 
10: claims data is.
 
11: BY MR. PENNEKAMP:
 
12: Q. Do you ask the providers for their
 
13: invoices for the medical services?
 
14: MR. MACEJKA: Object to form.
 
15: A. We will request for EOBs or billing
 
\vspace{2ex}
 
What is the exact text of just line 7 above?
 
\vspace{1ex}
 
\hrule
 
    \vspace{1ex}
 
The exact text of line 7 is: "Q. That's what you referred to a medical"
 
    \vspace{1ex}
 
    \hrule
 
    \vspace{1ex}

\caption{GPT-4o incorrectly answering a simple question about a page from a line-numbered witness deposition transcript. Line numbers, exactly as shown above, are passed to GPT-4o, which incorrectly answers 13\% of such one-page, 15-line deposition retrieval prompts.}
 
\vspace{-20pt}
 
    \label{fig:textat_gpt4_fail}
\end{figure}
 
However, no prior work has 
investigated LLMs' ability to handle the most very basic text tasks done in daily legal practice.   
In this paper, we introduce a benchmark for \textbf{B}asic \textbf{L}egal \textbf{T}ext (BLT) handling. BLT mimics simple tasks done by lawyers and paralegals, including looking up citations to the record,   finding statutory citations containing particular text or definitions, or finding the text in a contract that another lawyer has explicitly cited. 
 
BLT is designed so that a paralegal, new lawyer, or even non-subject matter expert could perform zero-shot at or near 100\%.  But we find that leading publicly available LLMs perform poorly out-of-the-box.  An example of this poor performance is in Figure~\ref{fig:textat_gpt4_fail}, where GPT-4o fails to retrieve the text on the specified line of a deposition transcript excerpt that is a mere 15 lines long.  

Yet we find that fine-tuning a less-advanced model, GPT-3.5-turbo, on BLT's training set leads to near-human (i.e., 100\%) performance. We also find that this fine-tuned model also performs better on a more complex legal task, demonstrating BLT's value for fine-tuning LLMs for legal applications.  
 
The BLT dataset has additional strengths.  Our code\footnote{ \url{https://github.com/blairstanek/blt}.} can generate unlimited new examples never before seen in corpora.  Moreover, BLT is scalable to different window sizes and is one of the few datasets taking full advantage of LLMs with window sizes of 64k or 128k tokens.  
 
\section{Background}
 
Law is a largely text-based profession and thus is often used to demonstrate LLMs' capabilities.  OpenAI's GPT-4 technical report \citep{openai2023gpt4} 
mentioned only a single benchmark in the abstract itself: GPT-4 had passed the bar exam, with the score in the 90th percentile.
 
OpenAI's developer livestream by co-founder Greg Brockman introducing GPT-4 \citep{Gpt4livestream} used four examples to show GPT-4's capabilities.  One involved U.S. tax law, where Brockman prompted GPT-4 with several tax-code sections and had it calculate the taxes of hypothetical taxpayers Alice and Bob.  Brockman proclaimed that GPT-4 can ``do taxes.’’


 
 
\subsection{Legal Use of LLMs}
 
Legal NLP is concerned with a diverse range of tasks, reflecting the diversity of tasks lawyers perform.  Examples include legal judgment prediction \citep{medvedeva-mcbride-2023-legal,chalkidis19neural,xiao18cail}, contract review \citep{cuad_hendrycks2021}, document review \citep{lenora_docreview}, and retrieving relevant case law \citep{kim22coliee}.  There has been extensive discussion of how NLP can benefit the legal system \citep{zhong2020does, legalsingularity}. LLMs have been deployed for a wide range of legal tasks, including case analysis \citep{savelka23can}, discovery \citep{pai-etal-2023-exploration}, and analyzing contracts \citep{roegiest-etal-2023-questions}.  

Several LLMs have been fine-tuned on legal materials \citep{colombo2024saullm7bpioneeringlargelanguage} and for legal tasks.  \citet{dominguezolmedo2024lawmapowerspecializationlegal} discovered that for some legal tasks, a 8-billion-parameter LLM that has been lightly pretrained on legal tasks substantially outperforms GPT-4, which has several orders of magnitude more parameters.  
 
Many legal benchmarks for LLMs have been created \citep{chalkidis22lexglue,fei2023lawbench}, with many incorporated into the broad LegalBench project  \citep{guha2022legalbench}.  These are all much higher-level tasks than BLT, including identifying testimony to which the hearsay doctrine applies or whether contractual terms impose particular restrictions \citep{cuad_hendrycks2021}. 
 
The SARA (StAtutory Reasoning Assessment) dataset \citep{holzenberger20dataset} is one of the higher-level tasks in LegalBench. 
It consists of nine tax-related sections of the U.S. Code, plus 376 hand-crafted ``cases’’ consisting of facts and a question that can be unambiguously answered applying the nine sections to the facts. Because SARA is a higher-level task clearly predicated on lower-level text handling (specifically, finding text at a citation), we use it to measure the effectiveness of fine-tuning with BLT, discussed in Section~\ref{sec:finetune}.

The ability of GPT-3 to handle SARA  was evaluated in \citet{gpt3statutory}, with lackluster performance found.  Qualitatively, GPT-3 often retrieved text from the wrong part of the given statute.  For example, GPT-3 was prompted with the text of the U.S. tax code's section 152, some facts about Alice and Bob, and the question of whether Alice's relationship to Bob fell under section 152(d)(2)(C).  GPT-3's response analyzed the question using the text of section 152(d)(2)(\textbf{D}), with the result that GPT-3 answered the question about Alice and Bob incorrectly.  This inability to retrieve clearly specified text -- resulting in incorrect answers to legal questions -- was a motivation behind the BLT dataset and this paper. 
 
LLMs have seen much of the internet during their training. To evaluate LLMs' ability to handle novel legal questions, \citet{nay2023large} generate synthetic multiple-choice legal questions. Similarly, to test whether LLMs can handle truly novel legal texts, \citet{gpt3statutory} generate synthetic sections constructed with nonces (phonetically plausible nonsense words) and probe \mbox{GPT-3's} ability to reason over these synthetic sections. BLT incorporates such synthetic sections, albeit for simpler tasks than statutory reasoning.
 
 
 
 

\subsection{Related LLM Evaluations}
 
Numerous evaluation metrics have been developed for LLMs \citep{Shahriar2024gpt4o,chang2023survey}. For example, BIG-Bench \citep{srivastava2023beyond} includes basic word handling tasks like word sorting and text editing.  Parsing software logs is evaluated by \citep{le2023evaluation}.  Simplifying complex sentences is evaluated in \citep{wu2024indepth}. LLMs can solve quite complicated tasks by being prompted to provide a chain of thought \citep{kojima2022large}, including in the legal domain \citep{yu22legal}. More generally, choosing the appropriate way to prompt LLMs, called prompt engineering, often has a substantial impact on LLM performance \citep{white2023prompt, survey_prompt}. A complementary approach has been to decompose the task at hand into tasks the LLM can handle \citep{dua22successive,khot22decomposed}. LLMs have been evaluated in various professional domains, including medicine \citep{beaulieu2024evaluating} and accounting \citep{zhao2024unleashing}.

LLMs have been trained or otherwise induced to use ``Tools’’ \citep{schick2023toolformer, paranjape2023art}.  For example, an LLM might detect that it needs to call a calculator tool to handle a math problem posed to it in text form.  In theory, tools could be written to handle the BLT tasks and then be integrated into LLMs.  But the BLT prompts are oversimplified versions of tasks lawyers need LLMs to do seamlessly.  A lawyer will not ask an LLM for the citation to the record where the plaintiff says ``I have therapy tomorrow.’’  But a lawyer might expect an LLM to insert a citation to the record that proves a plaintiff receives psychological care, and a basic text-matching tool would not handle that.
 
Some of BLT's prompts are quite long, which is realistic because lawyers regularly handle long texts.   \citet{liu2023lost} investigated how LLMs handle retrieval from long prompts.  They found that LLMs' accuracy followed a U-curve with respect to the information's position, with information in the middle of the prompt used much less than if it were at the start or end.  They connected this to the ``serial-position’’ effect exhibited by humans, who best remember material presented near the beginning or end. 
 
\section{The BLT Benchmark}
 
The BLT benchmark involves three different types of legal text, each of which has between two and five different tasks run on it.
 
\subsection{Deposition Transcripts}
In litigation in the U.S., depositions of witnesses under oath are a key factfinding tool. The depositions typically occur in lawyers' offices and allow lawyers to ask witnesses questions on virtually any topic.  Professional court reporters transcribe the depositions into transcripts, typically with 25 numbered lines per page, often running over 100 pages for a single witness deposition.  Attorneys must cite relevant portions of the resulting transcripts in subsequent motions, such as those asking the court to grant their side summary judgment.  Portions of transcripts are cited by page and line number. 
 
One basic legal text-handling task a lawyer must do is finding the page and line of a transcript where particular text appears.  This motivates the \textbf{text$\rightarrow$cite} task, where the prompt consists of one or more pages of actual deposition transcript followed by the question, ``What are the page number and line number of the line above with the text "\_\_"?’’ (after single pages, the prompt does not ask for the page number).  To ensure there is only one clearly correct answer, prompts are not constructed asking about lines with less than four words, that are subsets of another line, or that are too similar to other lines (defined as Levenshtein distance  under four \citep{levenshtein1966binary}).

The converse is another basic text-handling task:  given a citation to a transcript, find the text at the cited location.  Lawyers must do this basic task in order to evaluate the opposing side's motions.  Paralegals  do it on their side's own motions before submitting them \citep{ProParalegal}.  This motivates the \textbf{cite$\rightarrow$text} task, where the prompt consists of one or more pages of a deposition transcript followed by the question ``What is the exact text of just line \_\_ of page \_\_ above?’’ (after single pages, the prompt omits the page number).  An example of GPT-4o failing this task appears in Figure \ref{fig:textat_gpt4_fail}.
 
BLT's deposition transcript tests are built from a novel corpus we constructed of 33,176 lines of actual deposition transcripts filed with federal district courts and downloaded via Bloomberg Law from the Public Access to Court Electronic Records (PACER) system. They are from a variety of cases; our main criterion was that the transcripts be cleanly OCR'ed and could be fully cleaned with regular expressions.  This corpus can be extended \textit{ad infinitum} by downloading further transcripts and cleaning them, as  deposition transcripts are not copyrightable.\footnote{Lipman v. Massachusetts, 311 F. Supp. 593 (D. Mass. 1970); 1 Nimmer on Copyright \S 5.12[C]. Even in the unlikely event transcripts are copyrightable, this use of them is likely fair use. 17 U.S.C. \S 107.}  
The existing page and line numbers are stripped out, and random spans of appropriate length are selected, with new page numbers and line numbers, in case the LLM saw the original numbering during training. 
 
The size of the prompt is scalable, which was a key goal of BLT.  For BLT-4k (the version fitting within a 4,000 token window), we use a mix of 1 and 2 page subsets of depositions.  Larger versions have larger subsets. For BLT-128k (fitting inside a 128,000 token window), we use a mix of 120 and 140-page subsets of depositions.  Appendix~\ref{sec:transcript_sizes} details the sizes used.

\subsection{Synthetic Sections}
 
Lawyers regularly work with hierarchical text, including statutes, contracts, regulations, treaties, court rules, and corporate charters.  Hierarchical text is often organized into sections, subsections, paragraphs, subparagraphs, etc.  Being able to navigate such hierarchical text is a basic legal text-handling task required of all lawyers, whether they are litigators arguing that a statute applies to their case or are transactional lawyers negotiating contract terms.

\begin{figure}
    \hrule
    \vspace{1ex}
                Section 5217.  Definition of cleight.
                
                (a) General rule. The term "cleight" means any baitisolist or any roussiont.
                
                (b) The term "baitisolist" means any ballinated or any caset.
                
                (c) The term "roussiont" means any dicemercu or any accodpoileare.
                
                \vspace{2ex}
                
                What is the exact citation above where the term "roussiont" is defined? (Use standard legal formatting like section 1001(b)(2)).
                \vspace{1ex}
 
\hrule
    \vspace{1ex}
 
Section 5217(b)
 
    \vspace{1ex}
\hrule
    \vspace{1ex}
 
\caption{Example of GPT-4 incorrectly answering \textbf{defined$\rightarrow$cite} question with a 2-deep, 2-wide synthetic section. The correct answer is ``section 5217(c)’’.}
 
\vspace{-10pt}
 
\label{fig:defined_at_syn_gpt4_fail}
 
\end{figure}

We generate synthetic hierarchical sections, following the approach of \citet{gpt3statutory}. They use repeated application of the logical form $A \Rightarrow B$, with two parameters: depth, the maximum number of times it is applied; and width, the number of times it is applied to each $B$.  The terms defined are nonces that are not real words but are pronounceable.  These synthetic sections can be arbitrarily large, by adjusting the width and depth, thus making tasks based on them scalable to different-sized token windows.  For example, BLT-4k has synthetic sections ranging from 2-wide, 2-deep, as in Figure \ref{fig:defined_at_syn_gpt4_fail}, which are very short, up to 3-wide, 4-deep, which takes up much of the 4k token window.  At the highest end, BLT-128k has a variety ranging from 60-wide, 2-deep to 5-wide, 5-deep, which (because size is exponential with respect to depth) takes up much of the 128,000-token window.  For the full list of sizes in each \mbox{BLT-*}, see Appendix~\ref{sec:synsec_sizes}. Being synthetic ensures they are novel and not seen by LLMs during training.  This simulates the challenges faced by lawyers in handling newly drafted contracts, legislation, or other hierarchical text.  Nearly unlimited quantities of synthetic sections of any sizes can be generated by permuting the nonces. 
 
A basic legal text-processing skill is finding the citation, in a hierarchical text, of the text to which you are pointing a court or another lawyer.  This motivates applying the \textbf{text$\rightarrow$cite} task on synthetic sections, where the prompt consists of one synthetic section followed by the question ``What is the exact citation above of the text "\_\_"? (Use standard legal formatting like section 1001(b)(2)).’’  The code we use to generate synthetic sections guarantees there is only a single correct answer. 
 
The converse legal skill is, given a hierarchical citation, finding the text at it.  Hence we apply the \textbf{cite$\rightarrow$text} task to synthetic sections, with the prompt consisting of one synthetic section followed by the question ``What is the exact text of just section \_\_ above?’’ 
 
We ask this question only of ``leaves’’ in the statute, meaning they have no subsections underneath them.  This ensures there is only a single correct answer.  For example, suppose that section 573(a) was not a leaf, perhaps with paragraphs 573(a)(1) and 573(a)(2) underneath it.  If you asked for the text of section 573(a), it is ambiguous whether you should also return the text of 573(a)(1) and 573(a)(2).  Such ambiguity is avoided by considering only leaves.
 
We also include two other basic legal text-handling tasks on the synthetic sections.  Terms are defined in hierarchical texts and often referenced elsewhere in the same hierarchical text.  Lawyers must be able to cite a term's precise definition.  With \textbf{defined$\rightarrow$cite}, the prompt is one synthetic section followed by the question ``What is the exact citation above where the term "\_\_" is defined? (Use standard legal formatting like section 1001(b)(2)).’’  Conversely, when given such a citation by another lawyer, a lawyer must be able to find the term, which motivates \textbf{cite$\rightarrow$define}.  There, the prompt is one synthetic section followed by the question ``What is the term defined at section \_\_?’’  An example of GPT-4 incorrectly answering a defined$\rightarrow$cite problem appears~in~Figure \ref{fig:defined_at_syn_gpt4_fail}.
\subsection{U.S. Code}
 
The U.S. Code is the official compilation of general and permanent U.S. federal statutes.  The U.S. Code is a large corpus of hierarchical text.  We apply to the U.S. Code all four tasks that we applied to synthetic sections: \textbf{text$\rightarrow$cite}, \textbf{cite$\rightarrow$text}, \textbf{defined$\rightarrow$cite}, and \textbf{cite$\rightarrow$defined}.  For these four tasks on the U.S. Code, the prompt is the same as for synthetic sections.

During training, LLMs have doubtless seen all of the U.S. Code, which is not copyrighted and is publicly available on multiple websites.  To test whether LLMs' familiarity with U.S. Code sections causes errors, we add a fifth test for U.S. Code sections: \textbf{cite$\rightarrow$amended}.  In all but one respect, this test is identical to cite$\rightarrow$text, in that it has the text of one or more sections and asks ``What is the exact text of just section \_\_ above?’’ about a leaf.
 
The sole difference is that we make a small but semantically-important change to the text in that leaf to see if the LLM returns the original text or the changed text (which is the correct answer).  This tests a basic legal skill: applying a given newly-amended statute, rather than its old version.  If the leaf contains any numbers, we add or subtract one from the last appearing number (although we never move from 1 to 2 or from 2 to 1 since that would also require changing singular nouns to plural or vice versa). Otherwise, we tweak the last appearing citation from, say, ``(D)’’ to ``(A)’’. Otherwise, we toggle the last ``and’’ to ``or’’ or vice versa. Otherwise, we toggle the last ``shall’’ to ``may’’ or vice versa. If none of these changes are available, we insert ``unless otherwise provided by section 101,’’ at the start of the leaf. 
 
For all tasks on the U.S. Code, we do not use sections containing tables, which are not purely text.  Examples include the income tax tables at 26 U.S.C. \S 1, and 5,946 sections are excluded for this reason.  We do not use sections, like 5 U.S.C. \S 9507 and 25 U.S.C. \S 5329, with quoted hierarchical text such as model contracts, which are hard for even a human lawyer to read.  We never use any of the cites that Congress has sloppily added twice, such as the two subsection (e)'s in 42 U.S.C. \S 1397hh.
 
For text$\rightarrow$cite, we do the same test as with transcripts, not using lines that are under four words long, are subsets of any line appearing elsewhere in the prompt, or that have a Levenshtein distance under four from another line in the prompt.  For defined$\rightarrow$cite, we do not use terms defined in more than one place in the prompt. 

\begin{table*}[!ht]
\setlength{\tabcolsep}{4.7pt}
\begin{center}
\begin{tabular}{ l | l | c c| c c c c | c c c c c|c|}
 
&  & \multicolumn{2}{c |}{transc.}  & \multicolumn{4}{c|}{synthetic section} & \multicolumn{5}{c|}{U.S. Code}  &  \\
& \multicolumn{1}{c|}{model}  & \rotatebox[origin=b]{90}{text$\rightarrow$cite}
& \rotatebox[origin=b]{90}{cite$\rightarrow$text}
& \rotatebox[origin=b]{90}{text$\rightarrow$cite}
  & \rotatebox[origin=b]{90}{cite$\rightarrow$text}
  & \rotatebox[origin=b]{90}{defined$\rightarrow$cite}
  & \rotatebox[origin=b]{90}{cite$\rightarrow$defined}
& \rotatebox[origin=b]{90}{text$\rightarrow$cite}
  & \rotatebox[origin=b]{90}{cite$\rightarrow$text}
  & \rotatebox[origin=b]{90}{cite$\rightarrow$amended}
  & \rotatebox[origin=b]{90}{defined$\rightarrow$cite}
  & \rotatebox[origin=b]{90}{cite$\rightarrow$defined}  &    \multicolumn{1}{c|}{ \textbf{mean}} 
  \\
\hline
 
\multirow{7}{*}{BLT-4k}  & GPT-3.5-turbo & 53 & 32 & 72  & 38 & 83 & 79  & 89 & 52 & 56 \footnotesize{(0)}   & 77 & 98  & 66.3  \\  
  
 & GPT-4 & 82 & 78 & 88 &  97 & 90 & 100 & 98 & 93 & 93 \footnotesize{(0)} & 98 & 100    & 92.5 \\  
 & GPT-4-turbo & 87 & 88 & 85 & 63 & 76 & 95 & 98 & 84 & 77 \footnotesize{(7)} & 96 & 99 & 86.2  \\  

 
& chat-bison-32k & 84 & 29 & \phantom{0}7 & 77 & 37 & 92 & 83 & 90 & 89 \footnotesize{(1)} & 81 & 97 & 69.6 \\ 
 
& Claude-2.1 & 54 & 38 & 74 & 71 & 78 & 85 & 97 & 87 &87 \footnotesize{(1)} &96 & 95 & 78.4 \\ \cline{2-14}

 & GPT-4o & 88 & 84 & 99 & 90 & 98 & 98 & 99 & 91 & 94 \footnotesize{(0)} & 97 & 100 & 94.4 \\  

& Claude-3.5 & 96 & 80 & 100 & 99 & 100 & 100 & 100 & 97 & 97 \footnotesize{(0)} & 98 & 100 & 97.0 \\ 

\hline
\multirow{6}{*}{BLT-8k}  

  & GPT-4 & 44 & 26 & 64 & 49 & 82 & 83 & 94 & 74 & 76 \footnotesize{(0)} & 88 & 97 & 70.6  \\ 
 
  & GPT-4-turbo & 57 & 53 & 66 & 45 & 75 & 74 & 94 & 80 & 71 \footnotesize{(3)} & 98 & 99 & 73.8   \\ 

 
  & chat-bison-32k & 59 & \phantom{0}6 & \phantom{0}9 & 29 & 48 & 59 & 70 & 83 & 86 \footnotesize{(1)} & 80 & 89 & 56.2 \\ 
 
  & Claude-2.1 & 35 & 11 & 58 & 51 & 70 & 54 & 91 & 81 & 79 \footnotesize{(4)} & 94 & 92 & 65.1   \\ \cline{2-14} 
  
 & GPT-4o & 79 & 37 & 86 & 73 & 88 & 87 & 97 & 85 & 86 \footnotesize{(3)} & 99 & 96 & 83.0\\

  & Claude-3.5 & 75 & 66 & 94 & 72 & 99 & 99 & 100 & 100 & 94 \footnotesize{(1)} & 100 & 99 & 90.1  \\ 

\hline
 
\multirow{5}{*}{BLT-16k}   
 
& GPT-4-turbo & 30 & 20 & 78 & 36 & 83 & 83 & 90 & 64 & 58 \footnotesize{(5)} & 93 & 95  & 66.4  \\ 

& chat-bison-32k & 42 & \phantom{0}6 & 36 & 30 & 69 & 57 & 25 & 54 & 52 \footnotesize{(2)} & 43 & 58 & 42.9  \\ 
 
& Claude-2.1 & 21 & \phantom{0}5 & 64 & 48 & 76 & 61 & 82 & 66 & 65 \footnotesize{(4)} & 85 & 85 & 59.8 \\ \cline{2-14} 

& GPT-4o & 59 & 17 & 93 & 86 & 98 & 96 & 96 & 65 & 63 \footnotesize{(2)} & 93 & 96 & 78.4 \\ 

  & Claude-3.5 & 69 & 36 & 94 & 68 & 97 & 96 & 99 & 72 & 73 \footnotesize{(2)} & 94 & 95 & 78.4 \\
 
\hline
 
\multirow{5}{*}{BLT-32k} & GPT-4-turbo & 23 & 11 & 42 & \phantom{0}7 & 64 & 63 & 77 & 38 & 36 \footnotesize{(1)} & 82 & 85 & 48.0 \\ 

& chat-bison-32k & 13 & \phantom{0}0 & 8 & 10 & 39 & 32 & 12 & 32 & 36 \footnotesize{(0)} & 30 & 36 & 22.5 \\ 
 
& Claude-2.1 & 20 & \phantom{0}3 & 45 & 24 & 54 & 38 & 76 & 45 & 43 \footnotesize{(3)} & 71 & 70 & 44.5 \\  \cline{2-14} 

& GPT-4o & 50 & 17 & 82 & 64 & 83 & 76 & 96 & 48 & 44 \footnotesize{(6)} & 95 & 95 & 68.2 \\  

& Claude-3.5 & 72 & 35 & 62 & 33 & 82 & 72 & 90 & 57 & 56 \footnotesize{(6)} & 91 & 88 & 67.1 \\  

\hline
 
\multirow{3}{*}{BLT-64k} & GPT-4-turbo & 17 & \phantom{0}4 & 27 & 10 & 60 & 54 & 51 & 19 & 16 \footnotesize{(4)} & 55 & 66 & 34.5  \\ 

& Claude-2.1 & \phantom{0}5 & \phantom{0}1 & 52 & 26 & 56 & 35 & 58 & 31 & 29 \footnotesize{(4)} & 55 & 66 & 37.6 \\ \cline{2-14} 
 
 & GPT-4o & 65 & 16 & 87 & 61 & 92 & 88 & 95 & 53 & 53 \footnotesize{(1)} & 91 & 93 & 72.2   \\ 

\hline
 
\multirow{3}{*}{BLT-128k} & GPT-4-turbo & \phantom{0}9 & \phantom{0}3 & \phantom{0}3 & \phantom{0}0 & 20 & 17 & 34 & 12 & 12 \footnotesize{(0)} & 51 & 56 & 19.7 \\

& Claude-2.1 & 10 & \phantom{0}0 & 11 & \phantom{0}3 & 16 & 21 & 47 & 21 & 18 \footnotesize{(1)} & 45 & 60 & 22.9  \\ \cline{2-14} 

& GPT-4o & 53 & 10 & 36 & 18 & 72 & 49 & 79 & 30 & 28 \footnotesize{(1)} & 88 & 88 & 50.1 \\

\hline
 
\end{tabular}
\end{center}
 
\caption{Accuracy in percent of several models against all the different sizes of BLT.  GPT-4o and Claude-3.5 are broken out separately since their training cutoff was several months after the BLT dataset was made public online, so they may have seen the BLT test data during training. Under cite$\rightarrow$amended, the number in parentheses is how often the model erred by returning the unamended U.S. Code text rather than the amended text provided to the model in the prompt, results discussed in Subsection \ref{sec:cite2amended_problems}.}
\label{table:BLT_performance}
\end{table*}
 
Unlike synthetic sections, which can be generated in unlimited quantities in arbitrarily large sizes, there are a limited number of U.S. Code sections.  But it is a huge corpus, with 43,916 sections that meet the criteria discussed above, 447,037 leaves, and 23,562 unique definitions.  Although 94\% of sections are under 2,000 GPT-4 tokens, that still leaves 2,602 sections over 2,000 tokens, including 813 sections over 4,000 tokens and 196 sections over 8,000 tokens.  When there are insufficient numbers of large enough sections, we can generate prompts of any desired size \textit{ad infinitum}  by adding randomly selected other sections of approximately the same size.  We randomly shuffle the order of the sections in the prompt so that the target section's position is not a cue to the model.

Having multiple sections in a prompt resembles how OpenAI's Greg Brockman pasted nine tax-related sections 
into GPT-4 during the livestream introducing GPT-4.  This is realistic: lawyers handling real-world issues often must apply several statutes in conjunction, not just one.
 
\subsection{General Considerations}
 
For each of the 11 tests, and for each possible size (ranging from BLT-4k to BLT-128k), we generate a training/test split of 1000/100 prompts.\footnote{All BLT data can be downloaded from: 

\url{http://nlp.jhu.edu/law/blt/BLT-4k.zip}

\url{http://nlp.jhu.edu/law/blt/BLT-8k.zip}

\url{http://nlp.jhu.edu/law/blt/BLT-16k.zip}

\url{http://nlp.jhu.edu/law/blt/BLT-32k.zip}

\url{http://nlp.jhu.edu/law/blt/BLT-64k.zip}

\url{http://nlp.jhu.edu/law/blt/BLT-128k.zip}
.}  Why only 100 test prompts for each test split? Three reasons.  First, there are 11 tests, thus 1,100 test prompts for each size of BLT-*.  Second, the monetary cost of calling many LLMs with just 1,100 BLT-8k prompts with around 5,000 tokens per prompt is already nontrivial.  Third, any LLM deployed for real-world legal practice  should be at or near 100\%, and as accuracy approaches 100\% the t-statistic goes to~zero. 
 
 
Prompt engineering is not our focus, because the BLT tasks are expressly designed to be subsidiary tasks: legal users would not ask LLMs to solve the BLT tasks themselves. Rather, LLMs being able to handle the BLT tasks will generally be a prerequisite to accomplishing higher-level tasks, like drafting court documents citing statutes and deposition transcripts. 
Moreover, lawyers are unlikely to engage in more than rudimentary prompt engineering.  We do try different prompts, as discussed in section~\ref{sec:gpt4_trans_t2c} and appendices~\ref{sec:gpt4_trans_t2c_sensitivity} and \ref{sec:start_vs_end}.   
 
\section{Results and Discussion}
 
We tested four models from OpenAI:  GPT-3.5-turbo, GPT-4, and GPT-4-turbo, and GPT-4o.  From Google, we tested the chat-bison-32k variant of PaLM 2. From Anthropic, we tested Claude-2.1 and Claude-3.5 Sonnet.  (Due to token limits, we were not able to test Claude-3.5 on our two largest test sets, BLT-64k and -128k.) 

Crucially, GPT-4o and Claude-3.5's training cutoffs were several months after we released the BLT test and training data sets on the internet in November 2023.  They may have seen the BLT test sets during training, artificially boosting their performance.  Accordingly, they are broken out separately in Table~\ref{table:BLT_performance}.    
 
Table \ref{table:BLT_performance} contains our results.  All tests were by API call, with temperature set to 0.0 to maximize reproducibility and minimize hallucination.  Each number in Table 1 (other than the means and the numbers in parentheses) corresponds to 100 calls to the relevant LLM's API -- one call each for each of the 100 prompts in each test set. For example, there are 100 prompts in test for text$\rightarrow$cite for synthetic sections in BLT-4k. Our code measures accuracy with forgiving rules, ignoring case and whitespace.  Our code uses handwritten rules to classify errors, a feature we draw on in the discussion below. 

We note that accuracy monotonically decreases as prompt size increases, with each model achieving higher accuracy on BLT-4k than BLT-8k, higher accuracy on BLT-8k than BLT-16k, and so on.

\subsection{GPT-4 on transcript text\texorpdfstring{$\rightarrow$}{->}cite}
 
\label{sec:gpt4_trans_t2c}
 
All GPT-4 models (GPT-4, GPT-4-turbo, and GPT-4o) performed under 90 percent on transcript text$\rightarrow$cite for BLT-4k.  To further investigate this poor performance, we generate 1,000 new prompts in the same format (25-lines per page, with half being one-page and half being two-page) and pass them to GPT-4.  GPT-4 achieves 87.5\% on these 1,000.  Qualitatively, most errors are either identifying the line after the correct one or before the correct one.  The full error breakdowns are in Appendix~\ref{sec:gpt4_transcript_problems}.
 
The biggest determinant of performance is whether the transcript was a single page or two pages.  GPT-4 correctly answered  91\% of single-page transcript prompts, but just 84\% of 2-page transcript prompts.  This makes sense, since 2-page transcripts have 50 lines of text, whereas 1-page transcripts have just 25 lines of text.  (An example of GPT-4 getting a wrong answer on a 2-page transcript appears in Appendix~\ref{sec:gpt4_fail_2page}.)

To see whether the problem is the greater number of lines or having the text split into two pages, we generate 500 new prompts with single pages but with 50 lines per page.  (In other words, the transcript quotation is all one page, but with line numbers starting at "1:" and ending at "50:", followed by the question).  We find GPT-4 achieves 84.8\% accuracy, nearly identical to the two-page transcripts, indicating the problem is length, not being split into two pages. 
  
\begin{figure}
\includegraphics[width=0.99\linewidth]{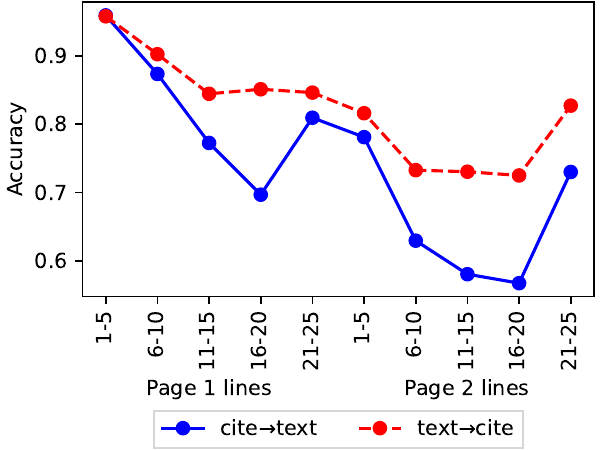}
 
\caption{Graph of location of relevant line versus accuracy on both transcript \textbf{cite$\rightarrow$text} and \textbf{text$\rightarrow$cite} on 5,000 prompts to GPT-4.}
 
\label{fig:graph_transcript_gpt4}
\end{figure}

To investigate how the location, within the transcripts, of the text impacts accuracy, we generate 5,000 new two-page prompts, and run against GPT-4.  The results are the red dashed line in Figure~\ref{fig:graph_transcript_gpt4}. We see a generally downwards trend. We do not see the distinct U-pattern observed by \citet{liu2023lost} in accuracy versus position of requested information within a long prompt. 

We also perform a sensitivity analysis, trying four question formats other than ``What are the page number and line number of the line above with the text "\_\_"?’’ appearing after the quotations.  (Details in Appendix \ref{sec:gpt4_trans_t2c_sensitivity}.) We find the biggest improvement simply by swapping the question from the end to the beginning and changing ``above’’ to ``below’’.  We tried the same switch -- moving the question from the bottom to the top -- for all of BLT-4k and re-ran against GPT-4.  The results are in Appendix~\ref{sec:start_vs_end}.  It turns out that moving the question from the end to the start actually hurt performance in 7 of 11 tasks, indicating no general trend. 
 
\subsection{GPT-4 on transcript cite\texorpdfstring{$\rightarrow$}{->}text}

All models performed poorly on transcript cite$\rightarrow$text for BLT-4k.  For example, Claude-3.5 got 80\%, GPT-4 got 78\%, and GPT-4o got 84\%.  To further investigate, we generated 1,000 new prompts in the same format and passed them to GPT-4, which got got 75.7\% accuracy on these 1,000.  We found little difference between one-page and two-page transcripts, on which GPT-4 got 76.6\% and 74.8\% respectively. 
 
To investigate how accuracy varies with the location of the requested cite, within the transcripts, we generate 5,000 new two-page prompts, and pass them to GPT-4.  The results are the solid blue line in Figure~\ref{fig:graph_transcript_gpt4}. We see a trend towards lower accuracy further into the transcript, with higher accuracy near the beginning and end of each page.

\subsection{Poor Performance on synthetic cite\texorpdfstring{$\rightarrow$}{->}text}
 
\begin{figure}
 
\includegraphics[width=0.99\linewidth]{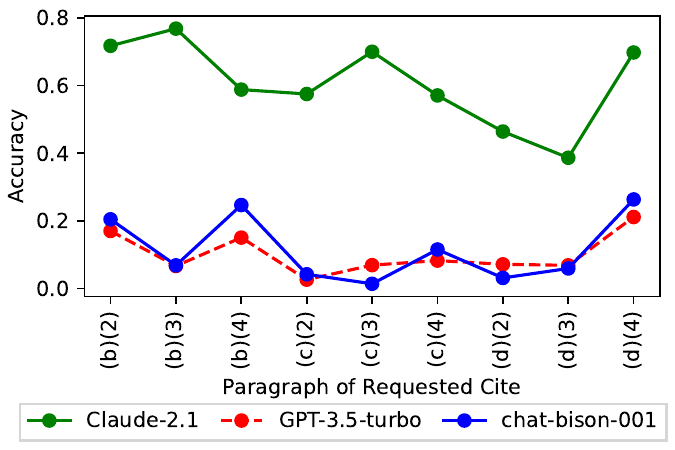}
 
\caption{Graph of location of requested cite versus accuracy for 5,000 synthetic \textbf{cite$\rightarrow$text} prompts, all using 3-wide, 4-deep synthetic sections, which are 127 lines long. Note that each first subdivision (e.g., (a), (1)) is used for a "General Rule" that has few lines, so such subdivisions are not included in this graph.}
\label{fig:graph_w3d4}
\end{figure}

Several models have their worst performance, among synthetic section tasks, on cite$\rightarrow$text.  Each BLT-* has a variety of different size sections, shown in Appendix \ref{sec:synsec_sizes}.  Some of the worst BLT-4k performance is on 3-wide, 4-deep synthetic sections.  To see if location within the section plays a role in accuracy, we generated 5,000 prompts using 3-wide, 4-deep synthetic sections, and we ran all against three models with poor performance on synthetic cite$\rightarrow$text.  The results are in Figure~\ref{fig:graph_w3d4}.  Once again, in contrast to \citet{liu2023lost}, there is no U-pattern.  We found the same lack of discernible pattern in accuracy versus location with all models and all tasks involving either synthetic sections or the U.S. Code.

 
 
\subsection{Problem Revealed by cite\texorpdfstring{$\rightarrow$}{->}amended}
 
\label{sec:cite2amended_problems}
 
Recall that for cite$\rightarrow$amended we make a minimal, but semantically-important, change to the subsection of the U.S. Code section being requested.  The correct answer is returning the subsection's text with this amendment.  Many of the errors involve returning the text of the wrong subsection.  But one type of error is particularly concerning: an LLM returning the subsection \textit{without} the amendment, presumably relying on the original U.S. Code text seen during training. 
 
In practice, this error means that, even if a lawyer or paralegal pastes in the new version of legislation, the LLM ignores it.  The LLM will act as if the legislation had never been amended. 
 
The occurrence of this error is in parentheses under the cite$\rightarrow$amended column in Table \ref{table:BLT_performance}, showing that several LLMs make this error a nontrivial number of times.  To investigate further, we generated 1,000 new BLT-4k style cite$\rightarrow$amended prompts.  On these, Claude-2.1 incorrectly returned the original text 17 times (1.7\%), chat-bison-32k did so 4 times (0.4\%), GPT-4o did so 10 times (1\%), and GPT-4-turbo did so fully
44 times (4.4\%). 
 
\section{Fine-Tuning}
\label{sec:finetune}

\begin{table}
\begin{center}
\begin{tabular}{l|c|c}
\multirow{2}{*}{text and task} & not & fine- \\
& tuned & tuned\\
\hline
transcript text$\rightarrow$cite & 53 & 100 \\
transcript cite$\rightarrow$text & 32 & 99 \\
synthetic text$\rightarrow$cite & 72 & 98 \\
  synthetic cite$\rightarrow$text & 38 & 100 \\
synthetic defined$\rightarrow$cite & 83 & 100 \\
synthetic cite$\rightarrow$defined & 79 & 100 \\
uscode text$\rightarrow$cite & 89 & 100 \\
uscode cite$\rightarrow$text & 52 & 100 \\
uscode cite$\rightarrow$amendedtext & 56& 100 \\
uscode defined$\rightarrow$cite &77 & 100\\
uscode cite$\rightarrow$defined & 98& 100 \\
 
\end{tabular}
\end{center}
 
\caption{Results of fine-tuning GPT-3.5-turbo on 9,900 training samples from BLT-4k.  Both numerical columns contain percent accuracy on BLT-4k's test prompts.  Fine-tuning GPT-3.5-turbo improves it to near perfect.  }
 
\label{table:fine_tune_success}
 
\end{table}
 
We fine-tune the 4,000-token version of GPT-3.5-turbo with BLT-4k's training set.  For each of the 11 task types, BLT-4k has a training set with 1,000 prompts and answers, for a total of 11,000 prompts and answers.  Recall that the training set and test set are generated in the same way, with the same code.  Of the 11,000 prompts and answers in the training set, we use 90\% for training, leaving 10\% as a possible evaluation set for future work.  We train for two epochs with all hyperparameters, like learning rate, set to the defaults.  The results of the fine-tuning are in Table~\ref{table:fine_tune_success}.  We find that fine-tuning brings GPT-3.5-turbo, which is far from OpenAI's most advanced model, to near the 100\% performance expected of lawyers and paralegals.

We tested how this fine-tuned GPT-3.5-turbo performs on the SARA dataset \citep{holzenberger20dataset}, using the 276 cases where the answer is entail/contradict.  Each prompt consists of each U.S. Code section(s) mentioned in the case, plus the facts (i.e., the premise) and the hypothesis (i.e., the question).  Without fine-tuning, GPT-3.5-turbo's accuracy was 54.3\% (150~/~276), but with our fine-tuning it rises to 60.9\%  (168~/~276).  Qualitatively, the fine-tuned model's answers focus on applying only the relevant statutory provisions. 
(An example is in Appendix \ref{sec:sara_example}.) 

This result shows the utility of fine-tuning LLMs on BLT's training data to improve performance on higher-level legal tasks. Also, this result is in line with the findings of \citet{dominguezolmedo2024lawmapowerspecializationlegal} that small LLMs that have been lightly fine-tuned on legal tasks often substantially outperform larger LLMs on legal tasks.  

\section{LLMs Trained After BLT Release}

We generally see the best performance from Claude-3.5 and GPT-4o, the two most recently released LLMs that we test.  There are three possible, non-exclusive explanations.  First, these models may be more advanced and thus better at most tasks.  Second, we released the BLT test sets on the internet in November 2023, several months before these models' training cutoff dates, so there may have been test-set leakage, with the models already having seen the answers during training.  Third, the BLT training sets (which are ten times larger than the test sets) were also released in November 2023, before these models' training cutoffs. As the fine-tuning experiments in Section \ref{sec:finetune} demonstrate, training on the BLT training data can substantially improve performance. Seeing the BLT training data may have improved these models' basic legal text handling.  

\section{Conclusion}
 
 
The chief innovation officer at a large international law firm observed to the \textit{New York Times} of LLMs, ``At its best, the technology seems like a very smart paralegal.’’ \citep{NYT}.  We find LLMs are more like sloppy paralegals.

Currently available LLMs perform poorly out-of-the-box on basic legal text handling.  The BLT tasks are designed to be truly basic, with humans able to perform them at or near 100\%.  The GPT-4 family, PaLM~2, and Claude-2.1 all fall far short.  Only one model, Claude-3.5, comes close, with 97.0\% aggregate performance, and then only on our smallest test set, BLT-4k.  Yet even Claude-3.5 achieves only 80\% on retrieving the text on a line of one or two pages of deposition transcript.  

We find that fine-tuning on our training set brings performance up to near 100\%.  We expect BLT to be a useful resource for those fine-tuning LLMs for much more complicated legal tasks, as well as a benchmark for LLMs' ability to do basic legal text handling without fine-tuning. 
 
 
 
\section*{Ethics Statement}
 
LLMs can be misused by legal professionals and laypersons alike to address legal problems properly requiring the full attention of a legal professional.  One of our goals is to alert potential users of the failings of existing LLMs at basic legal tasks.  Users may misconstrue our findings on the value of fine-tuning to assume incorrectly that an LLM with such fine-tuning can handle legal matters.  Even if an LLM gets 100\% on BLT, that does not mean the LLM can handle legal matters. 

\section*{Acknowledgments}

This work has been supported by the U.S. National Science Foundation under grant 2204926. Any opinions, findings, and conclusions or recommendations expressed in this article are those of the authors and do not necessarily reflect the views of the National Science Foundation. We would like to thank Susan McCarty and Orion Weller for valuable assistance.

\bibliography{acl_latex}

\clearpage

\appendix

\section{Synthetic Section Sizes}
\label{sec:synsec_sizes}
 
Larger versions of BLT have longer and more complicated prompts.  Below are the size of synthetic sections in each size of BLT.  The generated prompts are distributed uniformly among these section sizes.  For example, one-quarter of BLT-16k's synthetic statutes are 5-wide, 4-deep; one-quarter are 8-wide, 3-deep; and so on.
 
\vspace{2ex}
 
\begin{center}
   
\begin{tabular}{ l | l }
 
Version & Sizes \\
\hline
BLT-4k & 2 wide, 2 deep \\
& 2 wide, 3 deep \\
& 2 wide, 4 deep \\
& 2 wide, 5 deep \\
& 3 wide, 2 deep \\
& 3 wide, 3 deep \\
& 3 wide, 4 deep \\
& 4 wide, 2 deep \\
& 4 wide, 3 deep \\
\hline
BLT-8k & 2 wide, 6 deep \\
& 3 wide, 5 deep \\
& 4 wide, 4 deep \\
& 7 wide, 3 deep \\
& 20 wide, 2 deep \\
\hline
BLT-16k & 5 wide, 4 deep \\
& 8 wide, 3 deep \\
& 9 wide, 3 deep \\
& 30 wide, 2 deep \\
\hline
 
BLT-32k & 3 wide, 6 deep\\
& 4 wide, 5 deep\\
&6 wide, 4 deep\\
&12 wide, 3 deep\\
&11 wide, 3 deep\\
&44 wide, 2 deep\\
&40 wide, 2 deep\\
 
\hline
 
BLT-64k & 7 wide, 4 deep \\
&16 wide, 3 deep \\
&15 wide, 3 deep \\
&14 wide, 3 deep \\
&13 wide, 3 deep \\
&60 wide, 2 deep \\
&65 wide, 2 deep \\
 
\hline
 
BLT-128k & 4 wide, 6 deep \\
&5 wide, 5 deep \\
&8 wide, 4 deep \\
&9 wide, 4 deep \\
&20 wide, 3 deep \\
&80 wide, 2 deep \\
 
\end{tabular}
\end{center}

\section{Transcript Quotation Sizes}
\label{sec:transcript_sizes}
 
Larger versions of BLT have longer prompts.  Below are the number of pages of deposition transcript quotation used in each size BLT model. 
 
\vspace{2ex}
 
\begin{center}
\begin{tabular}{ l | l }
Version & Transcript Pages \\
\hline
BLT-4k & 1, 2 \\
BLT-8k & 5, 10, 15 \\
BLT-16k &   25, 40 \\
BLT-32k & 30, 60, 80 \\
BLT-64k & 100, 130 \\
BLT-128k & 120, 140 \\
\end{tabular}
\end{center}

\vspace{1ex}
 
The generated prompts are distributed uniformly among these page sizes.  For example, half of BLT-4k's prompts have one-page deposition quotations and the other half have two-page deposition quotations. 
 
\section{Example GPT-4 failing on a 2-page transcript text\texorpdfstring{$\rightarrow$}{->}cite Prompt}

\label{sec:gpt4_fail_2page}
 
\noindent Page 68:
 
\noindent 1: after the accident?
 
\noindent 2: A. He called me -- that same day,
 
\noindent 3: he called me at nighttime and asked me how
 
\noindent 4: I was.
 
\noindent 5: Q. He called you and asked you how
 
\noindent 6: you were?
 
\noindent 7: A. Yes.
 
\noindent 8: Q. What did you tell him?
 
\noindent 9: A. I just told him that I wasn't
 
\noindent 10: feeling good, that I was just laying in
 
\noindent 11: bed.
 
\noindent 12: Q. And what did he say?
 
\noindent 13: A. He said take the pills I gave
 
\noindent 14: you, you'll be ready and better by next
 
\noindent 15: week. He was just giving me something like
 
\noindent 16: a fast way for me to feel up, to cheer up.
 
\noindent 17: It wasn't helping.
 
\noindent 18: Q. Going back to your primary care
 
\noindent 19: physician. You said she wanted to send you
 
\noindent 20: to a specialist?
 
\noindent 21: A. Yes.
 
\noindent 22: Q. But then you called workers'
 
\noindent 23: comp?
 
\noindent 24: A. Yeah, that's when the issue
 
\noindent 25: happened and I called workers' comp. 
 
\vspace{2ex}

\noindent Page 69:
 
\noindent 1: Q. Then who did you treat with
 
\noindent 2: next?
 
\noindent 3: A. I treated with CityMD.
 
\noindent 4: Q. You said CityMD?
 
\noindent5: A. Yes.
 
\noindent6: Q. Was the place that you went to
 
\noindent7: the first time the same place that you went
 
\noindent8: like after you went to your primary care
 
\noindent9: physician?
 
\noindent10: A. Excuse me again?
 
\noindent 11: MR. GROSSMAN: You said you
 
\noindent 12: went to CityMD. She thinks you went
 
\noindent 13: twice, did you go twice?
 
\noindent 14: THE WITNESS: No.
 
\noindent 15: MR. GROSSMAN: Then you're not
 
\noindent 16: listening to the question. You went
 
\noindent 17: to CityMD first, then you went to
 
\noindent 18: your primary physician. Now, she
 
\noindent 19: wants to know where you went next.
 
\noindent 20: THE WITNESS: No, first I went
 
\noindent 21: to the primary.
 
\noindent 22: MR. GROSSMAN: First primary,
 
\noindent 23: and then CityMD and then where did
 
\noindent 24: you go?
 
\noindent 25: THE WITNESS: I was in CityMD
 
\vspace{2ex}
 
\noindent What are the page number and line number of the line above with the text "your primary physician. Now, she"?

\noindent\hrulefill
 
\noindent The page number is 69 and the line number \\
is 19.
 
\noindent\hrulefill

Note that GPT-4's answer above is wrong, since the requested text is on line 18, not 19.

\section{ GPT-4 Errors on Transcripts}
\label{sec:gpt4_transcript_problems}

GPT-4 performs surprisingly poorly on transcripts, even the 1- and 2-page transcripts in BLT-4k.  Below are GPT-4's errors on the large runs used to investigate the problem.  Our API-calling code automatically categorizes the errors. 
 
\subsection{GPT-4's errors on transcript text\texorpdfstring{$\rightarrow$}{->}cite from BLT-4k}
 
Recall that we generated 1,000 new BLT-4k style transcripts with text$\rightarrow$cite prompts to investigate GPT-4's poor performance.  Below is a tabulation of the lines returned by GPT-4 relative to the correct ones.  We see that the overwhelming majority of errors are returning the line before or line after the correct one.
 
\vspace{1ex}
\noindent \begin{center}
        \begin{tabular}{c|l}
\hline
\# & \textbf{Relative Error}   \\ \hline  
3 & prior page   \\
\hline
1 & 19 before  \\
\hline
1 & 3 before  \\
\hline
1 & 2 before  \\
\hline
12 & 1 before (i.e. prior line)  \\
\hline
76 & 1 after (i.e. next line)  \\
\hline
1 & 2 after  \\
\hline
1 & 3 after  \\
\hline
1 & next page  \\  \hline
    \end{tabular}
\end{center}
 
\vspace{2ex}
 
\subsection{GPT-4's  errors on transcript cite\texorpdfstring{$\rightarrow$}{->}text from BLT-4k}
 
\label{sec:qualitative_trans_c2t_errors}
 
Recall that we generated 1,000 new BLT-4k style transcripts with cite$\rightarrow$text prompts to investigate GPT-4's poor performance on that task.  Below is a tabulation of the errors. 
 
\vspace{1ex}
\noindent \begin{center}
        \begin{tabular}{c|l}
\hline 
\# & \textbf{Type of Error}   \\ \hline
166  & wrong line \\ \hline
    48  & superset  \\ \hline
    10  & not found  \\ \hline
    9  & subset wrong line  \\ \hline
    7  & multiple wrong lines  \\ \hline
    3  & subset  \\ \hline
    \end{tabular}
\end{center}
 
\vspace{2ex}
 
There was a difference in the type of errors on one-page versus two-page.  Almost all (84\%) the errors on two-page transcripts was returning the text from the wrong line (almost always the line before or after the correct one).  But one-page transcripts were only 51\% wrong-line errors, with 35\% being superset errors, where GPT-4 returned not only the text of the requested line, but also part of adjacent lines.  Here is an example of a superset error that occurred on this dataset:
\vspace{1ex}
 
...
 
\noindent 22: MS. WOLVERTON: Objection. Vague. Whose
 
\noindent 23: e-mail records?
 
\noindent 24: Q Ms. Abedin's e-mail records located. I
 
\noindent 25: apologize.
\vspace{1ex}
 
\noindent What is the exact text of just line 25 above?
\vspace{1ex}

\noindent GPT-4 returned ``Q Ms. Abedin's e-mail records located. I apologize.’’ which is actually lines 24 and 25, whereas the prompt asks for ``the exact text of \textbf{just} line 25 above’’.  These superset errors are much more understandable from a human perspective than an entirely wrong line, but they still fail according to the clear instructions in the prompt, with the word  `just’.
 
Of the 1,000 prompts, 166 were ``wrong line’’ errors.  Below shows the position, relative to the correct one, from which the text was incorrectly drawn:
\vspace{1ex}
\noindent \begin{center}
        \begin{tabular}{c|l}
\hline 
\# & \textbf{Relative Error}   \\  \hline
2 & prior page   \\  \hline
4 & same page, $>$3 lines before \\  \hline
1 & 3 lines before \\      \hline
6 & 2 lines before \\  \hline
81 & prior line \\  \hline
39 & next line \\  \hline
10 & 2 lines after \\  \hline
7 & 3 lines after \\  \hline
6 & 4 lines after \\  \hline
15 & same page, $>$4 lines after \\  \hline
3 & next page   \\ \hline

    \end{tabular}
\end{center}
 
\noindent As with text$\rightarrow$cite, here we see the overwhelming majority of wrong line errors are returning either the prior line or next line.
 
\section{Sensitivity analysis on GPT-4's transcript text\texorpdfstring{$\rightarrow$}{->}cite errors}
\label{sec:gpt4_trans_t2c_sensitivity}
 
Recall that GPT-4 performs poorly on text$\rightarrow$cite, so we attempted a sensitivity analysis, seeing how performance on the 100 test prompts for the task in BLT-4k changed with changes in the phrasing within the prompt.  The results are below.

\begin{center}
   
\begin{tabular}{p{5.5cm}|c}
\hline
 
BLT-4k default  \textbf{ What are the page number and line number of the line above with the text "\_\_"?} & 82/100 \\
\hline
   Move question from end to beginning, so question is \textbf{ What are the page number and line number of the line below with the text "\_\_"?}  & 99/100 \\
   \hline
  Keep question at end, adding ``exact’’ so question is \textbf{What are the exact page number and the exact line number of the line above with the text "\_\_"?}     & 84/100 \\
  \hline
    Keep question at end, adding ``precise’’ so question is \textbf{What are the precise page number and the precise line number of the line above with the text "\_\_"?}     & 84/100 \\
    \hline
Keep default question at end, but add the following introduction at the start: \textbf{Below is a portion of a transcript, with each line starting with a number that is important for referring to that line.} & 93/100 \\
\hline
   
\end{tabular}
\end{center}
\noindent The improvement from moving the question to the start motivated further experimentation, below.
 
\section{Question at Start versus End}
 
\label{sec:start_vs_end}
 
Because transcript text$\rightarrow$cite on GPT-4 saw large improvements from moving the question from the bottom (which is the standard for all of BLT-* for all tasks) to the top, we attempted the same change for all BLT-4k, running against GPT-4:
 
\begin{center}
\begin{tabular}{l|c|c|}
\multirow{2}{*}{text and task} & \multicolumn{2}{c|}{question at...} \\
 
& end & start \\
\hline
transcript text$\rightarrow$cite & 82 & 99 \\
transcript cite$\rightarrow$text & 78 & 85 \\
synthetic text$\rightarrow$cite & 88 & 91 \\
  synthetic cite$\rightarrow$text & 97 & 82 \\
synthetic defined$\rightarrow$cite & 90 & 82 \\
synthetic cite$\rightarrow$defined & 100 & 98 \\
uscode text$\rightarrow$cite & 98 & 96\\
uscode cite$\rightarrow$text & 93 & 82 \\
uscode cite$\rightarrow$amendedtext & 93 & 82 \\
uscode defined$\rightarrow$cite & 98 & 71\\
uscode cite$\rightarrow$defined & 100 & 100 \\
 
\end{tabular}
\end{center}
 
\noindent We see that GPT-4 is quite sensitive to whether the question is at the top or bottom.  But moving the question to the start actually produced worse results for 7 of the 11 tasks.  This indicates that the improvement in  transcript text$\rightarrow$cite by moving the question to the top was an outlier.

\section{Sensitivity analysis on GPT-4's transcript cite\texorpdfstring{$\rightarrow$}{->}text errors}
\label{sec:gpt4_trans_c2t_sensitivity}
 
Recall that GPT-4 did poorly on transcript text$\rightarrow$cite and that we performed a sensitivity analysis, discussed in Appendix \ref{sec:gpt4_trans_t2c_sensitivity}.  Since GPT-4 also did pooorly on cite$\rightarrow$text, we also do a sensitivity analysis on that. 
 
\begin{center}
   
\begin{tabular}{p{5.5cm}|c}
\hline
  BLT-4k default \textbf{What is the exact text of just line \_ of page \_ above?}    & 78/100 \\
  \hline
  Move question from end to beginning, so question is \textbf{What is the exact text of just line \_ of page \_ below?}   &  85/100\\
  \hline
  Rephase as \textbf{Return the exact text of just line 4 of page 59 above?} & 74/100 \\
\hline
Change ``exact’’ to ``precise’’, so question is \textbf{What is the precise text of just line \_ of page \_ above?} & 79/100 \\
\hline
Add clarification, so question is \textbf{What is the exact text of just line \_ of page \_ above? Return just the text on that line and return none of the text on the line before or after, even if necessary for a full sentence.} & 62/100 \\
\hline
Keep default question at end, but add the following introduction at the start: \textbf{Below is a portion of a transcript, with each line starting with a number that is important for referring to that line.} & 90/100 \\
\hline
 
\end{tabular}
\end{center}

\section{Example of fine-tuned GPT-3.5-turbo performing better}
 
\label{sec:sara_example}
 
Recall that fine-tuning GPT-3.5-turbo on part of BLT-4k's training set improved its performance on the SARA dataset \citep{holzenberger20dataset}.  Below we have an example SARA prompt used, plus both non-tuned GPT-3.5-turbo's answer and the fine-tuned version's answer.  We see the fine-tuned version sticks much more closely to the relevant statutory text and thus comes to the correct answer.
 
\vspace{1ex}
 
\noindent\hrule
\vspace{1ex}

\noindent \normalsize \textbf{The prompt (edited with ellipses for brevity):}
\small
\noindent We are going to be doing Entailment/Contradiction reasoning applying the statute(s) below:

\noindent§2. Definitions and special rules
 
\noindent(a) Definition of surviving spouse
 
\noindent \hspace{1ex}   (1) In general
 
\noindent \hspace{1ex}    For purposes of section 1, the term "surviving spouse" means a taxpayer-
 
\noindent \hspace{2ex}        (A) whose spouse died during either of the two years immediately preceding the taxable year, and
 
\noindent \hspace{2ex}        (B) who maintains as his home a household which constitutes for the taxable year the principal place of abode (as a member of such household) of a dependent (i) who (within the meaning of section 152) is a son, stepson, daughter, or stepdaughter of the taxpayer, and (ii) with respect to whom the taxpayer is entitled to a deduction for the taxable year under section 151.
 
***
 
\noindent (b) Definition of head of household
 
\noindent \hspace{1ex}     (1) In general
 ***
 
\noindent \hspace{1ex}      (2) Determination of status
 
\noindent \hspace{1ex}     Notwithstanding paragraph (1),
 
\noindent \hspace{2ex}          (A) an individual who is legally separated from his spouse under a decree of divorce or of separate maintenance shall not be considered as married;
 
\noindent \hspace{2ex}         (B) a taxpayer shall be considered as not married at the close of his taxable year if at any time during the taxable year his spouse is a nonresident alien; and
 
\noindent \hspace{2ex}         (C) a taxpayer shall be considered as married at the close of his taxable year if his spouse (other than a spouse described in subparagraph (B)) died during the taxable year.
 
***
 
\noindent Premise: Alice and Bob got married on Feb 3rd, 1992. Alice died on July 9th, 2014.
 
\noindent Hypothesis: Section 2(b)(2)(C) applies to Bob in 2014.
 
\noindent Answer:
 
\vspace{3ex}
 
\noindent\hrule
\vspace{1ex}

\noindent \normalsize \textbf{Non-tuned GPT-3.5-turbo's incorrect answer:}
\vspace{1ex}
 
\small
\noindent To determine whether Section 2(b)(2)(C) applies to Bob in 2014, we need to analyze the relevant provisions of the tax code.
 
\vspace{1ex}
\noindent Section 2(b)(2)(C) states that a taxpayer shall be considered as married at the close of his taxable year if his spouse (other than a spouse described in subparagraph (B)) died during the taxable year.
 
\vspace{1ex}
\noindent In this case, Alice, Bob's spouse, died on July 9th, 2014. To determine if Section 2(b)(2)(C) applies to Bob in 2014, we need to consider whether Alice's death during the taxable year qualifies Bob as a surviving spouse.
 
\vspace{1ex}
\noindent According to Section 2(a)(1)(A), a surviving spouse is defined as a taxpayer whose spouse died during either of the two years immediately preceding the taxable year. In this case, Alice died in 2014, which is within the two years immediately preceding the taxable year.
 
\vspace{1ex}
\noindent Next, we need to consider Section 2(a)(1)(B), which states that a surviving spouse must maintain as his home a household that constitutes the principal place of abode of a dependent. The dependent must be a son, stepson, daughter, or stepdaughter of the taxpayer, and the taxpayer must be entitled to a deduction for the taxable year under Section 151.
 
\vspace{1ex}
\noindent In this case, we don't have information about whether Bob has a dependent who meets the criteria mentioned above. Therefore, we cannot determine if Bob meets the requirements of a surviving spouse under Section 2(a)(1)(B).
 
\vspace{1ex}
\noindent Since we cannot determine if Bob meets the requirements of a surviving spouse, we cannot conclude whether Section 2(b)(2)(C) applies to Bob in 2014.
 
\vspace{3ex}
 
\noindent\hrule
\vspace{1ex}

\noindent \normalsize \textbf{Fine-tuned GPT-3.5-turbo's correct answer:}
 
\vspace{1ex}
 
\small
 
\noindent To determine if Section 2(b)(2)(C) applies to Bob in 2014, we need to consider the marital status of Bob and whether his spouse died during the taxable year.
 
\vspace{1ex}
\noindent According to the premise, Alice died on July 9th, 2014, and she was married to Bob. Therefore, we need to determine Bob's marital status at the close of the taxable year 2014.
 
\vspace{1ex}
\noindent Since Bob and Alice got married on Feb 3rd, 1992, and Alice died in 2014, Bob was married to Alice at the close of 2014.
 
\vspace{1ex}
\noindent Therefore, Section 2(b)(2)(C) applies to Bob in 2014 because his spouse (Alice) died during the taxable year.

\normalsize

\section{Dataset Statistics}

\begin{table*}[h!]
\setlength{\tabcolsep}{4.7pt}
\begin{center}
\begin{tabular}{ l | c c| c c c c | c c c c c|}
 & \multicolumn{2}{c |}{transc.}  & \multicolumn{4}{c|}{synthetic section} & \multicolumn{5}{c|}{U.S. Code}  \\
 
 & \rotatebox[origin=b]{90}{text$\rightarrow$cite}
& \rotatebox[origin=b]{90}{cite$\rightarrow$text}
& \rotatebox[origin=b]{90}{text$\rightarrow$cite}
  & \rotatebox[origin=b]{90}{cite$\rightarrow$text}
  & \rotatebox[origin=b]{90}{defined$\rightarrow$cite}
  & \rotatebox[origin=b]{90}{cite$\rightarrow$defined}
& \rotatebox[origin=b]{90}{text$\rightarrow$cite}
  & \rotatebox[origin=b]{90}{cite$\rightarrow$text}
  & \rotatebox[origin=b]{90}{cite$\rightarrow$amended}
  & \rotatebox[origin=b]{90}{defined$\rightarrow$cite}
  & \rotatebox[origin=b]{90}{cite$\rightarrow$defined} 
  \\
\hline

mean & 11.3 & 11.3 & 51.3  & 51.5 & 30.5 & 30.6 & 48.7 & 48.6 & 48.6 & 48.6 & 48.5 \\ 
  
stddev & 11.4 & 11.4 & 58.5 &  58.6 & 29.3 & 29.3 & 93.9 & 95.0 & 95.0 & 93.4 & 93.8 \\
median & 7 & 7 & 37 & 37 & 22 & 22 & 29 & 29 & 29 & 29 & 29 \\
min & 1 & 1 & 3 & 3 & 3 & 3 & 2 & 1 & 1 & 1 & 1 \\
max & 286 & 286 & 490 & 490 & 250 & 250 & 6383 & 6383 & 6383 & 6383 & 6383 \\
 
\hline

\end{tabular}
\end{center}

\caption{Number of words per sentence in the training set. Statistics computed across all sizes of the BLT dataset. Sentence and word boundaries were determined using NLTK's standard tokenizer \citep{bird09natural}.}
\label{table:BLT_stats_num_words}
\end{table*}

\begin{table*}[h!]
\setlength{\tabcolsep}{4.7pt}
\begin{center}
\begin{tabular}{ l | c c| c c c c | c c c c c|}  
 Adjective & 3 & 3 & 1 & 1 & 2 & 2 & 7 & 7 & 7 & 7 & 7 \\
 Adverb & 4 & 4 & 0 & 0 & 1 & 1 & 1 & 1 & 2 & 1 & 2 \\
 Conjunction & 11 & 11 & 2 & 2 & 3 & 4 & 19 & 19 & 19 & 19 & 19 \\
 Determiner & 7 & 8 & 15 & 15 & 25 & 23 & 11 & 11 & 11 & 11 & 11 \\
 Noun & 19 & 21 & 28 & 29 & 29 & 29 & 29 & 29 & 29 & 29 & 30 \\
 Number & 11 & 11 & 3 & 2 & 0 & 0 & 3 & 3 & 3 & 2 & 3 \\
 Pronoun & 8 & 6 & 0 & 0 & 0 & 0 & 0 & 0 & 0 & 0 & 0 \\
 Punctuation & 24 & 23 & 51 & 50 & 36 & 37 & 19 & 19 & 19 & 19 & 19 \\
 Verb & 13 & 13 & 0 & 0 & 3 & 4 & 10 & 10 & 10 & 10 & 10 \\
\hline
 
\end{tabular}
\end{center}
 
\caption{Part-of-speech tags in \% of occurrence, rounded to the closest percentage point. Statistics computed across all sizes of the BLT dataset, on a 1\% subset drawn at random from the training set for each task. Sentence and word boundaries were determined using NLTK's standard tokenizer \citep{bird09natural}. Part-of-speech tags were inferred using Stanford's POS tagger \citep{toutanova03feature}.}
\label{table:BLT_stats_pos_tags}
\end{table*}

We computed statistics on the BLT dataset using NLTK's tokenizers and Stanford's tagger \citep{bird09natural, toutanova03feature}. Number of sentences per document (Table~\ref{table:BLT_stats_num_sentences}) and number of words per sentence (Table~\ref{table:BLT_stats_num_words}) were computed on the entire training set. The distribution of part-of-speech tags (Table~\ref{table:BLT_stats_pos_tags}) was computed using 1\% of the training set, chosen at random. Given that these automatic tools were trained on standard written English, they may give somewhat inaccurate results on legal English.

\begin{table*}
\setlength{\tabcolsep}{4.7pt}
\begin{center}
\begin{tabular}{ l | l | c c| c c c c | c c c c c|}
 
BLT & & \multicolumn{2}{c |}{transc.}  & \multicolumn{4}{c|}{synthetic section} & \multicolumn{5}{c|}{U.S. Code}  \\
&  & \rotatebox[origin=b]{90}{text$\rightarrow$cite}
& \rotatebox[origin=b]{90}{cite$\rightarrow$text}
& \rotatebox[origin=b]{90}{text$\rightarrow$cite}
  & \rotatebox[origin=b]{90}{cite$\rightarrow$text}
  & \rotatebox[origin=b]{90}{defined$\rightarrow$cite}
  & \rotatebox[origin=b]{90}{cite$\rightarrow$defined}
& \rotatebox[origin=b]{90}{text$\rightarrow$cite}
  & \rotatebox[origin=b]{90}{cite$\rightarrow$text}
  & \rotatebox[origin=b]{90}{cite$\rightarrow$amended}
  & \rotatebox[origin=b]{90}{defined$\rightarrow$cite}
  & \rotatebox[origin=b]{90}{cite$\rightarrow$defined} 
  \\
\hline

\multirow{5}{*}{4k} & mean & 34.5 & 33.7 & 25.0 & 24.0 & 25.0 & 24.0 & 58.9 & 58.7 & 58.7 & 63.8 & 63.1  \\ 
 & stddev & 15.3 & 15.1 & 16.9 & 16.9 & 16.9 & 16.9 & 19.4 & 20.6 & 20.6 & 20.9 & 21.3 \\
& median & 32 & 31 & 21 & 20 & 21 & 20 & 58 & 58 & 58 & 63 & 62  \\
 & min & 8 & 8 & 8 & 7 & 8 & 7 & 5 & 4 & 4 & 9 & 8  \\
 & max & 81 & 79 & 57 & 56 & 57 & 56 & 119 & 219 & 219 & 133 & 132  \\
 
\hline

\multirow{5}{*}{8k} & mean & 215.4 & 214.6 & 93.6 & 92.6 & 93.6 & 92.6 & 121.4 & 119.7 & 119.7 & 127.1 & 127.4  \\ 
  
& stddev & 105.9 & 105.8 & 46.0 & 46.0 & 46.0 & 46.0 & 34.9 & 32.9 & 32.9 & 36.6 & 35.8 \\
& median & 205.5 & 205 & 98 & 97 & 98 & 97 & 124 & 122 & 122 & 127 & 129  \\
 & min & 58 & 57 & 26 & 25 & 26 & 25 & 34 & 29 & 29 & 23 & 32  \\
 & max & 472 & 471 & 165 & 164 & 165 & 164 & 213 & 210 & 210 & 240 & 238  \\
 
\hline

\multirow{5}{*}{16k} & mean & 722.4 & 721.6 & 104.5 & 103.5 & 104.5 & 103.5 & 231.7 & 231.7 & 231.7 & 233.5 & 231.5  \\ 
   & stddev & 262.7 & 262.5 & 56.0 & 56.0 & 56.0 & 56.0 & 62.3 & 62.0 & 62.0 & 65.0 & 65.0 \\
 & median & 644.5 & 643 & 95.5 & 94.5 & 95.5 & 94.5 & 231 & 231 & 231 & 240.5 & 235  \\
  & min & 315 & 315 & 36 & 35 & 36 & 35 & 68 & 61 & 61 & 90 & 89  \\
  & max & 1187 & 1186 & 191 & 190 & 191 & 190 & 435 & 401 & 401 & 389 & 390  \\
 
\hline

\multirow{5}{*}{32k} & mean & 1263 & 1263 & 234.7 & 233.7 & 234.7 & 233.7 & 461.8 & 460.4 & 460.4 & 462.6 & 458.0  \\ 
& stddev & 566.3 & 566.2 & 164.2 & 164.2 & 164.2 & 164.2 & 104.8 & 102.8 & 102.8 & 104.9 & 107.5 \\
& median & 1183 & 1181 & 174 & 173 & 174 & 173 & 469 & 473 & 473 & 478 & 470  \\
& min & 391 & 390 & 46 & 45 & 46 & 45 & 173 & 183 & 183 & 171 & 185  \\
& max & 2307 & 2305 & 489 & 488 & 489 & 488 & 768 & 767 & 767 & 740 & 752  \\
 
\hline

 \multirow{5}{*}{64k} & mean & 2531 & 2530 & 226.2 & 225.2 & 226.2 & 225.2 & 873.1 & 874.4 & 874.4 & 884.9 & 884.1  \\ 
  & stddev & 799.3 & 799.0 & 126.2 & 126.2 & 126.2 & 126.2 & 199.4 & 199.2 & 199.2 & 177.3 & 176.2 \\
 & median & 2721 & 2720 & 230 & 229 & 230 & 229 & 875 & 874 & 874 & 883 & 882 \\
  & min & 1365 & 1365 & 66 & 65 & 66 & 65 & 425 & 402 & 402 & 425 & 424  \\
  & max & 3648 & 3646 & 461 & 460 & 461 & 460 & 1377 & 1375 & 1375 & 1318 & 1378  \\
 
\hline

 \multirow{5}{*}{128k} & mean & 2874 & 2873 & 793.3 & 792.3 & 793.3 & 792.3 & 1817 & 1820 & 1820 & 1805 & 1818  \\ 
& stddev & 872.6 & 872.4 & 503.2 & 503.2 & 503.2 & 503.2 & 320.7 & 304.3 & 304.3 & 301.3 & 307.2 \\
 & median & 3306 & 3305 & 788.5 & 787.5 & 788.5 & 787.5 & 1804 & 1825 & 1825 & 1794 & 1816 \\
 & min & 1671 & 1671 & 86 & 85 & 86 & 85 & 904 & 903 & 903 & 983 & 923  \\
 & max & 3929 & 3926 & 1710 & 1709 & 1710 & 1709 & 2752 & 2657 & 2657 & 2776 & 2751  \\
 
\hline

\end{tabular}
\end{center}
 
\caption{Number of sentences per document in the training set. At most 4 significant digits were kept. Sentence boundaries were determined using NLTK's standard tokenizer \citep{bird09natural}.}
\label{table:BLT_stats_num_sentences}
\end{table*}

\end{document}